\pdfoutput=1
\documentclass[11pt]{article}
\usepackage{acl}
\usepackage{times}
\usepackage{latexsym}
\usepackage[T1]{fontenc}
\usepackage[utf8]{inputenc}
\usepackage{microtype}
\usepackage{inconsolata}
\usepackage{graphicx}
\usepackage{amsmath}
\usepackage{amsfonts}
\usepackage{booktabs}
\usepackage{multicol}
\usepackage{verbatim}
\usepackage{paralist}
\usepackage[toc,page]{appendix}
\usepackage{tabularx}
\title{Controlled Diversity: Length-optimized Natural Language Generation}

\author{Diana Marie Schenke \and Timo Baumann\\
Faculty of Informatics and Mathematics, OTH Regensburg\\
  \texttt{diana.schenke@st.oth-regensburg.de} 
  \texttt{timo.baumann@oth-regensburg.de} \\}

\begin{document}
\maketitle
\begin{abstract}
LLMs are not generally able to adjust the length of their outputs based on strict length requirements, a capability that would improve their usefulness in applications that require adherence to diverse user and system requirements.
We present an approach to train LLMs to acquire this capability by augmenting existing data and applying existing fine-tuning techniques, which we compare based on the trained models' adherence to the length requirement and overall response quality relative to the baseline model.
Our results demonstrate that these techniques can be successfully applied to train LLMs to adhere to length requirements, with the trained models generating texts which better align to the length requirements. Our results indicate that our method may change the response quality when using training data that was not generated by the baseline model. This allows simultaneous alignment to another training objective in certain scenarios, but is undesirable otherwise. Training on a dataset containing the model's own responses eliminates this issue.

\end{abstract}

\section{Introduction}

Generating texts while under strict length restrictions is a complex task that is relevant both in content creation and human-machine communication applications. Modern Large Language Models (LLMs) have been shown to possess the ability to create a wide range of content, which makes them useful in many different fields \cite[pp.\,70-77]{LLMSurvey}. The nature of LLMs however makes the exact length of the texts they generate unpredictable, making them ill-suited for applications with strict length requirements. In this paper, we present a method that enables LLMs to be useful for these kinds of tasks, a problem that has not been addressed in any published research paper.

The broader goal of fine-tuning LLMs to better adhere to a set of preferences, on the other hand, is a well researched field and has brought forth a whole range of different fine-tuning approaches \cite[pp.\,36-43]{LLMSurvey}. Besides utilizing the fundamental approach of supervised fine-tuning  (SFT), we adapt three reinforcement learning methods, namely Proximal Policy Optimization (PPO; \citealp{PPO}), Direct Preference Optimization (DPO; \citealp{DPO}) and Odds Ratio Preference Optimization (ORPO; \citealp{ORPO}) to achieve our objective. These methods are originally combined with the reinforcement learning with human feedback (RLHF) \cite{PPOFinetuning}
approach, in which the model learns from human evaluations. However, they have also previously been used to optimize automatically measured performance metrics \citep[e.\,g.][]{penzkofer-baumann-2024-evaluating}.
The difference between the generated and required length of output can also be automatically measured. Thus, we likewise perform fully automated RL.

Importantly, we generate the necessary training data through a data augmentation process that can easily be performed on most datasets. This makes it possible to add the length objective as a side objective, while the LLM also learns its central tasks (e.\,g.\ from human feedback). One example application of the integration of our approach into LLM fine-tuning is text simplification \citep{sentenceSimplification}, where simplified summaries of a user-specified length might further improve accessibility to complex content.

Other tasks have diversity requirements based on strict outside factors, creating length requirements that are user or situation specific, like a time limit based on user attention span or their commute time, or a space limit based on user eyesight, device settings or hardware properties. Spoken communication between a software assistant and a car driver is one example of a task that has to account for complex environmental factors, as the assistant's responses should never distract the driver. \citet{CarConversations} have demonstrated how non LLM based methods can improve the drivers' attention by carefully tuning response timing and length. To instead employ the more powerful and broad capabilities of LLMs for this task, exact control over response length is a necessity.

\section{Background}

LLMs are already being applied to a wide range of applications: they excel in classic NLP tasks like text generation and information extraction, are capable of performing information retrieval, act as recommender systems and evaluate both human and LLM generated content \cite[pp.\,70-77]{LLMSurvey}. A key element to their success is their ability to be adapted to very specific tasks, which is achieved by fine-tuning a pretrained LLM using various approaches. In the following sections, we will discuss some of these approaches and explore how they can be applied to enable LLMs to adhere to length requirements.

The simplest approach to adapt a pretrained LLM to a specific task is supervised fine-tuning (SFT), which relies on a relatively small dataset of demonstrations, that is often created by humans specifically for this purpose and is generally of high quality. The model learns by generating completions of partial data samples, using the actual completions in the dataset to generate a loss based on their difference. While models adapted using supervised fine-tuning generally perform better at these specific tasks than baseline models, supervised fine-tuning is often just the first step in a process called reinforcement learning with human feedback (RLHF), which is a widely used approach for adapting pretrained LLMs to new tasks. 

\citet{HumanFeedbackTraining} summarize the process of aligning an LLM to human feedback using reinforcement learning in the following three-step procedure:
\begin{compactenum}
    \item \textbf{Supervised Fine-Tuning:} This step consists of fine-tuning a pretrained LLM on a set of human curated demonstration data of prompts and desired responses.
    \item \textbf{Reward Model Training:} In this step, a reward model is trained to evaluate the outputs of the trained model from step one based on human provided preference data. 
    \item \textbf{Policy Optimization with Reinforcement Learning:} In the last step, the LLM is trained using a reinforcement learning algorithm, usually PPO \cite{PPO}. The reward model is used to generate training rewards in this step.
\end{compactenum}
During the third step, the expected reward for the responses generated by a policy \(\pi\) based on a set of prompts \(\mathcal{D}\) are simply the predictions of the reward model \(r(x,y)\) \cite{PPOFinetuning} for the response $y$ given prompt $x$: 
\begin{equation}
 \mathbb{E}_{x\sim\mathcal{D}, y\sim\pi(\cdot|x)} [r(x,y)] 
\end{equation}
Based on this, the following PPO objective can be formulated \cite{PPOFinetuning,DPO}:
\begin{equation} \label{eq:rlhf_ppo_objective}
\begin{split}
 \underset{\theta}{\text{maximize}} \quad &\mathbb{E}_{x\sim\mathcal{D}, y\sim\pi_{\theta_{\text{SFT}}}(\cdot|x)} [r(x,y)] \\ & - \beta \text{KL} \left[ \pi_{\theta_{\text{SFT}}}(\cdot|s_t), \pi_{\theta}(\cdot|s_t)\right]  
\end{split}
\end{equation}
where \(\theta_{\text{SFT}}\) refers to the model parameters after the supervised fine-tuning step, meaning that the model derived after this step is used as reference. 

Our objective is to minimize the difference between the actual length of the generated text and the target length, which can be measured automatically. Thus, some objective measure based on the length difference can be used as a reward, like squared difference, rendering the training of a reward model obsolete.

We thus define our reward function $r(y)$ for a given response $y$ as the squared difference between the length of the response $\text{len}(y)$ and the length target specified in the prompt $\text{len}_{\text{target}}$:
\begin{equation}
 r(y) = \left( \text{len}(y) - \text{len}_{\text{target}}  \right)^2
\end{equation}

The Direct Preference Optimization approach to LLM alignment, which was introduced by \citet{DPO}, eliminates the need for a reward model by instead using preference data to align the LLM. Preference data refers to a dataset containing two (or more) possible responses to a prompt, where one of the responses is labelled as ``preferred''. The goal of DPO is, on a high level, to train the LLM to generate outputs that are closer to the preferred responses in the data set and less like the responses that are not preferred.

\citet{DPO} reparameterize the PPO objective and apply the Bradley-Terry model to produce a reward based on the model's likelihood to produce the preferred and not preferred responses \(y_w\) and \(y_l\). They derive the following objective:
\begin{equation}
\begin{split} 
\mathbb{E}_{(x,y_w,y_l)\sim\mathcal{D}} \biggr[ \sigma \biggr( &\beta\log\frac{\pi_{\theta}(y_w | x)}{\pi_{\theta_{\text{SFT}}}(y_w | x)} \\ &-\beta\log\frac{\pi_{\theta}(y_l | x)}{\pi_{\theta_{\text{SFT}}}(y_l | x)} \biggr) \biggr]
\end{split}
\end{equation}
where the prompt \(x\), and the responses \(y_w\) and \(y_l\) are drawn from some preference dataset \(\mathcal{D}\). \citet{DPO} simply use the negative of this objective as the loss function \(\mathcal{L}_{\text{DPO}}(\pi_{\theta};\pi_{\theta_{\text{SFT}}} ) \).

Odds Ratio Preference Optimization (ORPO) is a fine-tuning approach proposed by \citet{ORPO} which aims to improve DPO by further streamlining LLM adaption through two main improvements: 

Firstly, they eliminate the need for the preference model found in DPO by computing their loss directly based on the odds that the model outputs an unwanted response over a preferred one. Specifically, they compute the odds ratio \(\text{OR}_\theta(y_w,y_l)\), which measures how many times more likely the model with the parameters \(\theta\) is to output the preferred response \(y_w\) over the unwanted response \(y_l\):
\begin{equation}
\text{OR}_\theta(y_w,y_l)  =  \frac{\text{odds}_\theta(y_w|x)}{\text{odds}_\theta(y_l|x)}
\end{equation}
where \(\text{odds}_\theta(y|x)=\frac{P_\theta(y|x)}{1-P_\theta(y|x)}\) is the probability of the model to generate \(y\) rather than not generating it (i.\,e., generating something else).

Based on this, they formulate the odds ratio loss:
\begin{equation}
    \mathcal{L}_{\text{OR}} = -\log\sigma\left(\log\frac{\text{odds}_\theta(y_w|x)}{\text{odds}_\theta(y_l|x)} \right)
\end{equation}
which is then combined with an SFT-style cross entropy loss \(\mathcal{L}_{\text{SFT}}\) that compares the token distribution generated by the model with the tokens in the reference data:
\begin{equation} 
\mathcal{L}_{\text{ORPO}} =  \mathbb{E}_{(x,y_w,y_l)\sim\mathcal{D}} \left[ \mathcal{L}_{\text{SFT}} + \lambda  \mathcal{L}_{\text{OR}}\right]
\end{equation}
This forms their second innovation, as it allows ORPO to directly adapt models without first performing the SFT step necessary in the PPO and DPO approaches to RLHF.

\begin{figure}[t]
  \centering
  \includegraphics[width=0.85\columnwidth]{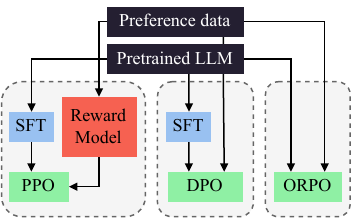}
  \caption{Process Diagram for the PPO, DPO and ORPO approaches to Reinforcement Learning with Human Feedback. All approaches require a pretrained LLM and a preference dataset, typically human-made.}
  \label{fig_RLHF_process_chart}
\end{figure}

Figure \ref{fig_RLHF_process_chart} outlines the way in which DPO and ORPO simplify the RLHF process. As previously explained, a reward model is not necessary for the training objective of adhering to length requirements, which is the scenario tested in our experiments. In a hypothetical multi-objective scenario, where adherence to length requirements is a side objective, a reward model may still be needed.

It is also important to point out here that making or adapting preference data for the objective of adhering to length requirements is quite simple, as the prompts can simply be augmented to include a length requirement that matches the length of the preferred response.

In terms of training outcomes, the papers introducing DPO \cite{DPO} and ORPO \cite{ORPO} both present their methods as superior to previous ones, while later research differs from those conclusions, often finding PPO superior to DPO and ORPO, and suggesting that DPO and ORPO might be limited in ways that PPO is not \cite[][pp.\,21-22, 29]{DPOPPOComparison,RLHFsurvey}. We conclude that there is no singular ``correct'' approach to RLHF and test all three approaches.

\section{Methods}

\begin{figure*}[t]
  \centering
  \includegraphics[width=0.9\linewidth]{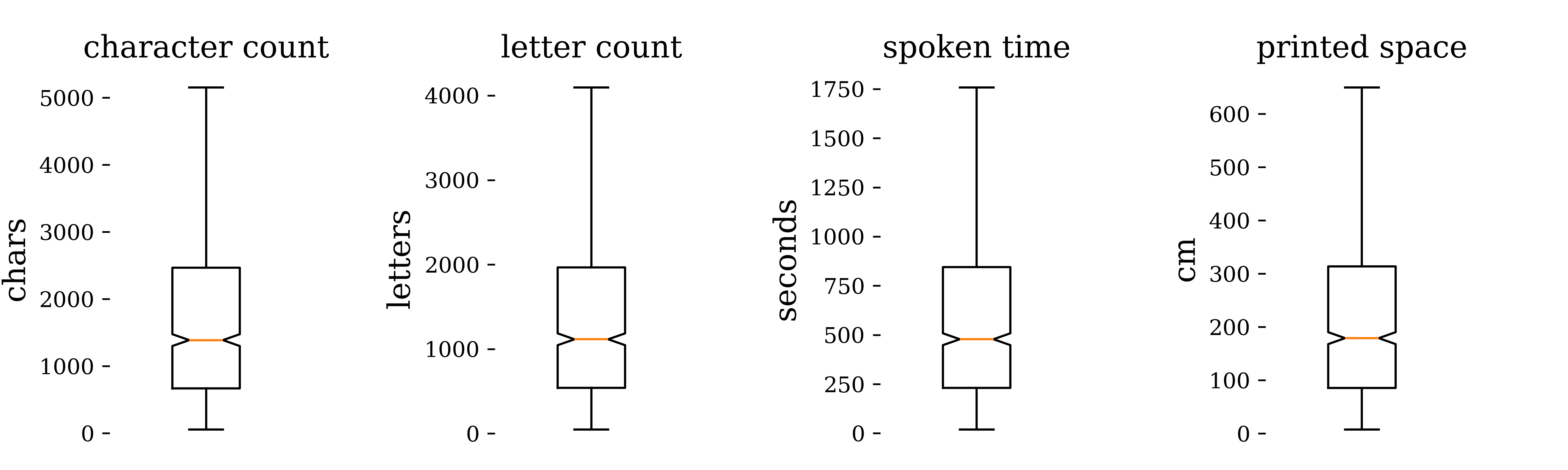}
  \caption{Distribution of length requirements in our test data set ($n=1280$ samples). Each of the four portions of the test data is of equal size.}
  \label{fig_data_len}
\end{figure*}

We derive a first dataset from the UltraChat dataset\footnote{We choose the UltraChat dataset for its large size and diverse content, which allows us to evaluate the fine-tuning process for multiple tasks and to reduce the risk of underfitting. A difference in data quality between synthetic (LLM-generated) and human-made data was not a primary concern, as we were not interested in improving the quality of the models responses. We specifically use samples from the ``questions about the world'' portion of UltraChat.}, which was collected and published by \citet{UltraChat}. This dataset is composed of around one million chat-style conversations between two instances of ChatGPT.

We use the first question-response pair of each conversation only and we augment the question by adding a short sentence stating the length requirement to its end. The length target in the augmented prompt is set to match the length of the response that is contained in UltraChat, making them the desired response. This method of data augmentation can be applied to any data set that is formatted in a prompt-response style, making it easy to integrate into an existing training setup for a given task.

We also create a second dataset containing the same questions as in UltraChat but responses generated by baseline Llama 3.1. This dataset has the disadvantage of being more time-consuming to create, but might offer two significant advantages: Firstly, training on responses created by the model ensures that the model doesn't alter its overall response behaviour during training by adapting to other data characteristics that aren't the length requirement. This is only relevant if adherence to length requirements is the sole objective, as simultaneously training the model to respond differently is desirable otherwise. 
Secondly, it allows us to test whether generating training data from scratch using only a prompt dataset is a viable approach, offering an alternative if using responses from an existing dataset impacts the response quality of the trained model.

We use the following length requirements in our experiments:

\begin{compactenum}
	\item \textbf{Characters:} Total number of characters (including white space and newlines) in the response.
	\item \textbf{Letters:} Total number of alphanumeric characters in the response (i.\,e.\ excluding punctuation and white space).
	\item \textbf{Speech:} Time in seconds it takes to utter the response, as estimated by \citet{SayIt} based on the characters in the text.
	\item \textbf{Print:} Horizontal length of text in centimeters when printed in 12pt Times New Roman without line breaks.
\end{compactenum}

The length requirement distributions for the test portion of our dataset are presented in Figure \ref{fig_data_len}.

We use the Llama 3.1 8b model for our experiments, which is part of a project by Meta AI to provide large state-of-the-art pretrained LLMs to the research community free of charge \citep{LlamaPaper}. Their newest models, which were released in 2024 and are referred to as the "Llama 3 herd of models", are a range of LLMs of different size and capability, which were developed by \citet{Llama3Paper}. We additionally use Quantized Low Rank Adaption (QLora, \citealp{QLoRA}) during model adaptation to shorten the training time and reduce the memory usage of the model.

To measure if the response quality of the model degrades during training, we employ the following measures: To estimate response quality, we use the semantic similarity of the model's responses to high quality reference responses. We measure semantic similarity using the F1 measure of SemScore~\cite{SemScore} and use the ChatGPT-generated responses contained in the UltraChat dataset alongside the responses of the untrained baseline model as references.

As an additional precaution, we use the language tool utility \cite{languageToolPython} to check for grammar, spelling and syntax errors.

We train Llama 3.1 using supervised fine-tuning for 10 epochs, each consisting of 128,000 samples from our first dataset, saving a copy of the model after every epoch. These copies are then compared on an evaluation dataset consisting of 1,280 new samples, with the model that performs best given its training time being chosen for further fine-tuning.

Choosing such a high number of epochs allows us to explore whether training for multiple epochs could be a useful strategy in a data limited scenario. We are specifically interested in examining whether there would be an inverse relation between the models' adherence to length requirements and the quality of its outputs, making such a strategy potentially disadvantageous. 

We then further train the best model from the previous experiment using PPO, DPO and ORPO\footnote{While ORPO is usually applied without any prior SFT training, doing so would have required the creation of another large preference data set containing baseline Llama 3.1 responses. We decided against creating this dataset due to resource constraints, and instead apply ORPO similarly to DPO, allowing us to use the same preference dataset for both.} to compare their relative performance in this application. We train DPO and ORPO on 32,000 samples and PPO on 8,000 samples\footnote{PPO training was much slower per sample than DPO and ORPO in our experiments.}.

\begin{figure}[t]
  \centering
  \includegraphics[width=\columnwidth]{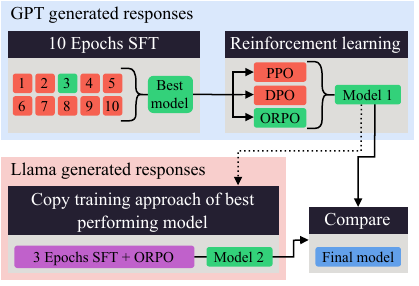}
  \caption{A schematic overview of our training process, described in detail in the Methods section. The most successful models in each step are highlighted in green.}
  \label{fig_traingdiagram}
\end{figure}

We then compare the performance of the resulting models both with each other and the baseline SFT-trained model, to find out whether any of the reinforcement learning based fine-tuning methods provide a significant enough performance increase to justify the additional training time.

Lastly, we train a second model on our second dataset containing Llama-generated responses using the most successful approach found above. Afterwards, we evaluate both models' response quality to test for the aforementioned concerns regarding bias.

A schematic overview of the whole training process is presented in Figure \ref{fig_traingdiagram}.

\section{Results}

\begin{figure*}[t]
  \centering
  \includegraphics[width=0.85\linewidth]{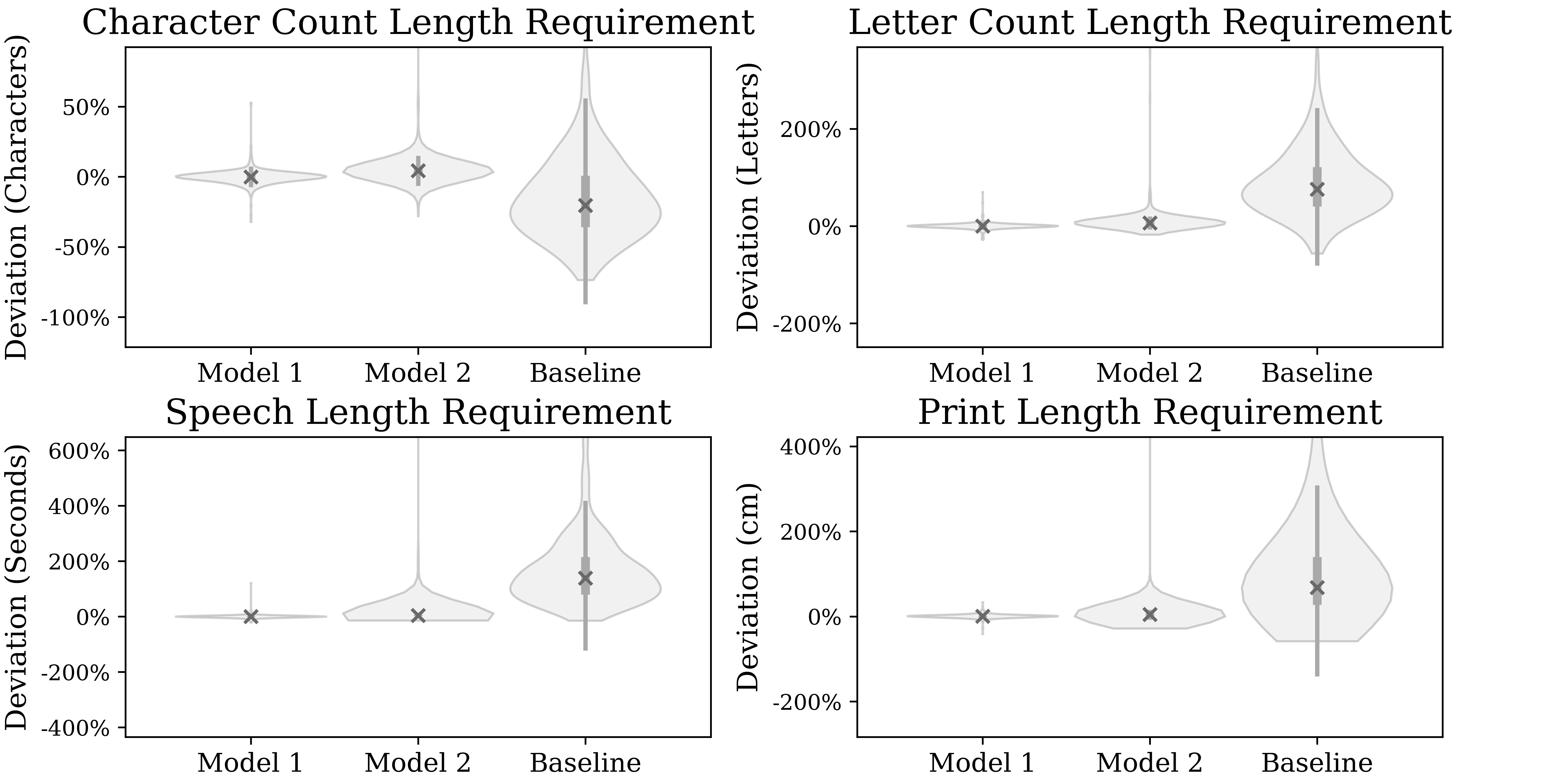}
  \caption{Distribution of percentage deviation from the length target for our two final models compared to the unoptimized baseline across four different types of length requirement. Note the different y-axes in the plots.}
  \label{fig_model_comp}
\end{figure*}

After training with supervised fine-tuning, the models responses align significantly closer to the length requirements. The mean relative deviation from the length requirement was reduced from $108\%$ to $7.61\%$ after one epoch ($\approx92\%$ decrease) and then to $6.05\%$ after the third (another $\approx21\%$ decrease). After the third epoch, the improvement became smaller and less stable, decreasing to $5.35\%$ after the tenth epoch (another $\approx12\%$ decrease), but also increasing intermittently.

We decide to pick the model that was trained for three epochs as a viable starting point for further experiments and comparisons as we felt improvements afterwards were too marginal to justify the additional training time.

The results of further fine-tuning the SFT model using reinforcement learning are mixed. The ORPO-trained model is the only one to consistently outperform the SFT model it is based on across all length requirements ($3.12\%$ mean relative deviation from length requirement, $\approx48.4\%$ decrease), even if the gain is only marginal on the character count length requirement. 
The DPO model's performance is inconsistent, outperforming the SFT model overall ($4.64\%$ mean relative deviation from length requirement, $\approx23.3\%$ decrease), but performs significantly worse on the character count and speech length requirements than both the ORPO model and the SFT baseline model.
The results of the PPO model are worse than those of the baseline SFT model ($7.16\%$ mean relative deviation from length requirement, $\approx18.3\%$ increase).

We therefore decide to base our further experiments on the ORPO model, as we deem its improvement big enough to be worth the additional training time. This model will be referred to as Model 1 below.
We train another model (Model 2) using the dataset that contains responses generated by the baseline Llama 3.1 model, using the same approach of three epochs of SFT followed by ORPO for comparability.

Of the two final models, Model 1 outperforms Model 2, deviating less from the length requirement (see Figure \ref{fig_model_comp} for details). Importantly, however, Model 2's responses share the same semantic similarity to the reference responses as the baseline model, indicating no deterioration in response quality after optimizing for length. 

In contrast, Model 1 responses are more similar to ChatGPT-generated responses, which is hardly surprising, given that the ChatGPT style was part of its training.
Overall, we do not find relevant changes in SemScore or gramaticality ratings, indicating that optimizing for length requirements does not lead to response deterioration.\footnote{Also, see Appendix~\ref{appendix:sample_responses} for some examples.}

With respect to the four kinds of length requirements, we find that these are, of course, highly correlated. Our trained models' performance (relative error of length) does not significantly differ between the length requirements, indicating that they all work similarly well. (However, models fail to generalize to another requirement, number of words in the response.)

\section{Discussion}
We presented a method for adapting LLMs to adhere to length requirements, demonstrating that existing fine-tuning techniques are sufficient for this task. Specifically, SFT and ORPO proved to be the most effective among the methods tested. 

The reason for this might lie in their design: SFT trains the model by adjusting its output directly based on the difference of individual output tokens compared to tokens in reference data, while PPO and DPO use a reinforcement learning-based approach where they propagate backwards from a reward that evaluates the entire output text\footnote{The reward is generated implicitly in the case of ORPO and DPO.}. ORPO combines both approaches. On a more abstract level, SFT-based approaches train the model to ``imitate'' the reference texts, while reinforcement learning approaches instead train it to produce texts that are evaluated highly based on an objective \cite[p.\,41]{LLMSurvey}. 

It has been demonstrated that this allows reinforcement learning-based approaches to be more successful than SFT when aligning LLMs to abstract objectives that center around learning from human feedback \cite{HumanFeedbackTraining,PPOforLLM,PPOFinetuning}. If adhering to length requirements is, however, a fundamentally simpler objective than other common fine-tuning objectives, then using SFT-based approaches might be sufficient, explaining the success of SFT in our experiments. 

The comparative under-performance of some reinforcement learning-based approaches could be explained by their inherent instability \citep[p.\,43]{LLMSurvey}, which might lead to longer convergence times. ORPO on the other hand, combining both approaches, might be able to utilize their respective advantages for further gains.

Overall, we find the following data augmentation approaches useful: when training for adherence to length requirements as a side objective, or when strict adherence to them is the most important concern, modifying the prompts in an existing prompt-response style dataset to include length requirements is both effective and efficient. However, note that this approach changes the response behaviour and therefore potentially the response quality of the LLM. In situations where this is unacceptable, using prompts from a dataset to generate responses using the baseline model, then augmenting the prompts with fitting length requirements, could be a better approach, but produces a model that tends to deviate more from the  requirement.

Limitations to our findings are the fact that we only tested one model that was trained on augmented data taken from a single dataset, oriented towards a single task. The length requirements we used were also quite long (see Figure \ref{fig_data_len}) on average, making it hard to generalize our findings to tasks that require very short responses.
In particular, we find that our models are unable to deal with length requirements outside the range that they were trained with, as is exemplified in Appendix~\ref{appendix:sample_responses}.

\bibliography{main}
\newpage

\begin{appendix}
\onecolumn
\section{Sample Responses of the Trained Model}\label{appendix:sample_responses}

Below, we show some sample responses of the model to the same prompt with varying length requirements. Note how the model fails to perform well for short lengths, which were not part of its training. Also, note the radical variability of air-speeds reported (25--100 mph, even in the same response). Such aspects are not well represented in semantic similarity metrics such as SemScore.

\begin{table}[h]
    \caption{Responses of Model 1 to the task \\\texttt{What is the air-speed velocity of an unladen swallow?\\ Generate precisely \emph{LEN} characters in your response.}
    \label{tab:example}}
    \centering
    \begin{tabularx}{\linewidth}{rXrr}
    \toprule
        LEN & response & actual & error \\ \hline
         10 & As an AI language model, I don't have personal opinions, but I can answer. & 74 & +640\,\%\\
         50 & The air-speed velocity of an unladen swallow is approximately 80 miles per hour (130 kilometers per hour). & 106 & +112\,\%\\[22pt]
         100 & The air-speed velocity of an unladen swallow is approximately 30 miles per hour (48 kilometers per hour). & 105 & +5\,\% \\
         150 & As an AI language model, I don't have the ability to answer this question as it is a joke. The air-speed velocity of an unladen swallow is not measurable. & 154 & +3\,\% \\
         200 & As an AI language model, I do not have the ability to determine the exact air-speed velocity of an unladen swallow. However, it is estimated to be around 80-90 mph (130-145 km/h). & 179 & -10\,\%\\
         250 & The air-speed velocity of an unladen swallow is approximately 80-100 mph (130-160 km/h). However, this can vary depending on the species of swallow and the conditions of the environment. Some species of swallows can reach speeds of up to 25 mph. & 245 & -2\,\%\\
         300 & The air-speed velocity of an unladen swallow is approximately 70-80 miles per hour (mph). However, this can vary depending on the type of swallow and its altitude. For example, the European swallow has a top speed of around 35-45 mph (55-72 km/h), while the American swallow can reach speeds of up to 35 mph (56 km/h). & 318 & +6\,\%\\
    \bottomrule
    \end{tabularx}
\end{table}

\end{appendix}
\end{document}